\documentclass[sigconf]{acmart}
\usepackage{subfigure}

\usepackage{adjustbox}
\AtBeginDocument{%
  \providecommand\BibTeX{{%
    \normalfont B\kern-0.5em{\scshape i\kern-0.25em b}\kern-0.8em\TeX}}}


\copyrightyear{2021}
\acmYear{2021}
\setcopyright{acmlicensed}\acmConference[CIKM '21]{Proceedings of the 30th ACM International Conference on Information and Knowledge Management}{November 1--5, 2021}{Virtual Event, QLD, Australia}
\acmBooktitle{Proceedings of the 30th ACM International Conference on Information and Knowledge Management (CIKM '21), November 1--5, 2021, Virtual Event, QLD, Australia}
\acmPrice{15.00}
\acmDOI{10.1145/3459637.3482130}
\acmISBN{978-1-4503-8446-9/21/11}

\begin{document}
\fancyhead{}
% \title{AMTL: Adaptively-Gated Twins-of-Router Layer For Effective and Efficient Embedding Learning}
\title{Learning Effective and Efficient Embedding via an Adaptively-Masked Twins-based Layer}

\author{
 Bencheng Yan$^{*}$, Pengjie Wang$^{*}$, Kai Zhang, Wei Lin, Kuang-Chih Lee, Jian Xu and Bo Zheng$^{\dagger}$
 }
  \affiliation{%
  \institution{Alibaba Group}
  % \country{China}
}
 \email{{bencheng.ybc,pengjie.wpj,victorlanger.zk,kuang-chih.lee,xiyu.xj,bozheng}@alibaba-inc.com, lwsaviola@163.com}
 \thanks{$*$ These authors contributed equally to this work and are co-first authors.}
 \thanks{$\dagger$ Corresponding author}

% \author{anonymous}


\begin{abstract}
Embedding learning for categorical features is crucial for the deep learning-based recommendation models (DLRMs). 
Each feature value is mapped to an embedding vector via an embedding learning process.
Conventional methods configure a fixed and uniform embedding size to all feature values from the same feature field. 
However, such a configuration is not only sub-optimal for embedding learning but also memory costly.
Existing methods that attempt to resolve these problems, either rule-based or neural architecture search (NAS)-based, need extensive efforts on the human design or network training. 
They are also not flexible in embedding size selection or in warm-start-based applications. 
In this paper, we propose a novel and effective embedding size selection scheme. 
Specifically, we design an Adaptively-Masked Twins-based Layer (AMTL) behind the standard embedding layer. 
AMTL generates a mask vector to mask the undesired dimensions for each embedding vector. 
The mask vector brings flexibility in selecting the dimensions and the proposed layer can be easily added to either untrained or trained DLRMs. 
Extensive experimental evaluations show that the proposed scheme outperforms competitive baselines on all the benchmark tasks, and is also memory-efficient, saving 60\% memory usage without compromising any performance metrics.
\end{abstract}




\maketitle

\section{Introduction}
\label{sec:Introduction}

% 图, efficiency (time cost), user id

% Click-through rate (CTR) prediction is an critical task in many applications such as recommendation systems, web search and online advertising, where the task is to predict the probability a user will click the items.
% Recently embedding learning based deep models (DLRM) have been successfully applied in various applications (e.g., recommendation systems) \cite{cheng2016wide,guo2017deepfm,wang2017deep} due to the ability for feature learning and the capacity for relation modeling (see Fig \ref{figure:An example of DLRM} as an example).
% One of the main part of these DLRM is the embedding layer which is used to learn the categorical features (examples of categorical features could include User ID, item ID and query ID).
% A standard way to exploit this categorical information in DLRM is to map each feature value to an embedding space \cite{guo2017deepfm,cheng2016wide,wang2017deep}.
% Specifically, given a set of categorical feature values $F$ in one feature field and its vocabulary size is $|F|$, each feature value $f_i$ in $F$ is mapped to a dense embedding vector by an embedding matrix $W^{|F|*D}$, where $D$ is the predefined embedding dimension for this feature field.
Recently, deep learning-based recommendation models (DLRMs) have been widely adopted in many web-scale applications such as recommender systems \cite{cheng2016wide,guo2017deepfm,wang2017deep,xdeepfm,li2019graph,juan2016field}. 
One of the main parts of DLRMs is the embedding layer, which exploits the categorical features.
 % (see Fig \ref{figure:An example of DLRM} as an example). 
A standard embedding layer maps the categorical feature to an embedding space \cite{guo2017deepfm,cheng2016wide,wang2017deep}. 
Specifically, given a feature field $F$ and let its vocabulary size be $|F|$, each feature value $f_i \in F$ is mapped to an embedding vector by an embedding matrix  $W \in \mathbb{R}^{|F|\times D}$, where $D$ is a predefined embedding dimension.

However, the above standard method can lead to two problems. 
First,  in real applications, different feature values in the same feature field can have significantly different frequencies. 
% For example, some users are more active than others in a recommender system. 
% Therefore, feature values in the feature field User ID can have quite different frequencies in the user behavior datasets. 
For high-frequency feature values, 
% a sufficiently large embedding dimension is usually necessary to well express them. 
it is necessary to use a sufficiently large embedding dimension to express rich information.
Meanwhile, assigning too large embedding dimensions to low-frequency feature values is prone to over-fitting issues. 
Therefore, a fixed and uniform embedding dimension for all the feature values in a feature field can undermine effective embedding learning for different feature values. 
Second, storing the embedding matrix with a fixed and uniform dimension may result in a huge memory cost \cite{zhang2020model,shi2020compositional,zhao2020memory}.
A flexible dimension assignment is needed to reduce the memory cost.

\begin{figure}[t]
\centering
\includegraphics[width = .4\textwidth]{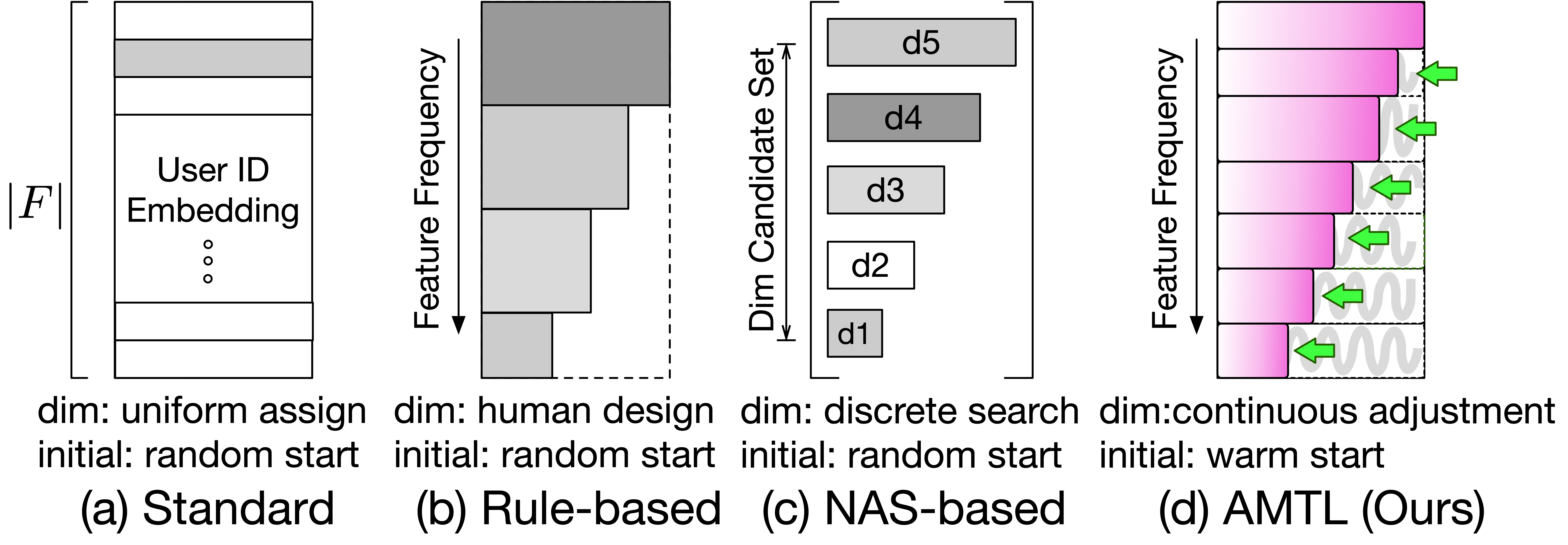}
\vspace{-1em}
\caption{Comparison among existing methods and ours.
}
\vspace{-2em}
\label{figure:Comparison among existing methods and ours}

\end{figure}

There are some existing works trying to learn unfixed and nonuniform embedding dimensions for different feature values. 
They can be primarily divided into two categories. 
(1) \emph{Rule-based methods} adopt human-defined rules, typically according to the feature frequencies, to give different embedding dimensions to different feature values \cite{ginart2019mixed}  (see Fig. \ref{figure:Comparison among existing methods and ours} (b) for an example). 
The problem with this category of methods is that they heavily rely on human knowledge and human labor. 
The resulting rough dimension selection for groups of feature values can often lead to poor performance (see Section \ref{sec:Click Prediction Task}). 
(2) \emph{Neural architecture search (NAS)-based methods} use NAS techniques to search from several candidate embedding dimensions to find a suitable one for each feature value \cite{zhao2020memory,zhao2020autoemb,joglekar2020neural,liu2020automated} (see Fig \ref{figure:Comparison among existing methods and ours} (c) for an example). 
These methods require careful design of the search space and training-searching strategies.
The search space is usually limited to a restricted set of discrete dimensions. 
% They can hardly give a satisfactory performance or a significant memory reduction (see Section \ref{sec:Click Prediction Task} and \ref{sec:Memory Cost Comparison}).
Besides, both categories of methods mentioned above require training (i.e., embedding learning) from scratch. 
However, in real applications, there may exist some embedding matrices already trained with a huge amount of data. 
Such embedding matrices can be utilized for warm starting (see Section \ref{sec:Effectiveness for the Warm Start of the Embedding Matrix}). 
Unfortunately, existing methods are not friendly to accommodate such a warm start mechanism.

% In this paper, we propose a new and effective scheme about embedding dimension selection which does not need too much human efforts or the help of NAS.
% The general idea is to apply a gated router layer behind the embedding layer.
% The gated router layer can adaptively learn a mask vector for each feature value to make a decision on embedding dimension, where this vector can mask the redundant embedding dimension to meet the requirement for various dimensions of different feature values.
% Specifically, we present a Adaptively-Gated Router Layer (AML) (Section \ref{sec:Adaptively-Gated Router Layer}) to generate the mask vector and further extend AML to the Adaptively-Gated Twins-Of-Router Layer (AMTL) to address imbalanced samples problem (see Section \ref{sec:Adaptively-Gated Twins-Of-Router Layer} for details).
% Then the masked embedding vectors can be taken as the embeddings with adaptive dimension, and be feed into DLRM directly.
% In this way, the proposed AML or AMTL can be joint trained with DLRM in an end-to-end manner.
% Such idea can help model leave away from the shortcomings of rule-based or NAS-based methods and bring a flexibility to the models.
% On the one hand, the embedding dimension can be adjusted in continuous dimensions (from the minimal dimension (one dimension) to the maximal dimension) rather than several predefined candidate dimensions.
% On the other hand, the proposed layer can be easily applied on any existing trained embedding matrices in DLRM and is friendly to the warm start of the embedding matrices.

In this paper, we propose a novel and effective method to select proper embedding dimensions for different feature values. 
The basic idea is to add an Adaptively-Masked Twins-based Layer (AMTL) on top of the embedding layer. 
Such a layer can adaptively learn a mask vector 
% for each feature value and the mask vector is applied to the embedding vector of that feature value. 
to mask the undesired dimension of the embedding vector for each feature value.
The masked embedding vectors can be taken as the vectors with adaptive dimensions and are fed into the subsequent processes in DLRMs. 
% More specifically, we design the Adaptively-Gated Twin-Of-Router (AMTL) layer  to generate the mask vectors (see Section \ref{sec:Adaptively-Gated Twins-Of-Router Layer}).
% AMTL can be jointly trained within the DLRM training process in an end-to-end manner. 
This method exhibits some nice properties. 
% First, the embedding dimension of different feature values can be learned with sufficient flexibility without human interaction or specific NAS design, which improves the model performance and saves the model memory.
% Second, the proposed gated router layer can be easily applied on top of any already trained embedding matrix and therefore is friendly to accommodate warm start mechanisms.
First, it is effective for embedding learning because the embedding dimension of different feature values can be learned and adjusted in continuous integer space with sufficient flexibility without human interaction or specific NAS design (see Section \ref{sec:Click Prediction Task}).
Second, it is efficient since a memory-efficient model can be built by adjusting the embedding dimension (see Section \ref{sec:Memory Cost Comparison}). 
Third, the parameters of the embedding matrix can be efficiently trained with the warm start mechanism (see Section \ref{sec:Effectiveness for the Warm Start of the Embedding Matrix}).

We summarize our contributions as follows:
% \begin{itemize}
% \item 
(1) We propose a novel embedding dimension selection method that completely removes the necessity of human rules or NAS architectures to facilitate adaptive dimension learning.
% \item 
(2) The proposed method (AMTL) can be easily applied in trained DLRMs to facilitate a warm start. 
The twins-based architecture successfully tackles the sample unbalance problem.
% \item 
(3) Extensive experimental results demonstrate that the proposed method outperforms strong baseline methods. 
The nice properties of AMTL helped us reduce memory cost by up-to 60\% without compromising any performance metrics, and can further improve the performance by the warm start mechanism.

\section{Method}
% In this section, we introduce the proposed method AMTL.
% First, we introduce the basic idea about our model (Section \ref{sec:Basic Idea}).
% Then we present the base version AML designing gated router layer to adaptively select dimensions for different feature values (see Section \ref{sec:Adaptively-Gated Router Layer})
% Finally, we further extend AML to AMTL to tackle the unbalance problems about samples (see Section \ref{sec:Adaptively-Gated Twins-Of-Router Layer}).
% Note the summarized nations are presented in Table \ref{table:Notation}.

\subsection{Basic Idea}
\label{sec:Basic Idea}
% To present the basic idea of our method, 
We first recall a standard embedding layer which can be expressed as
% \begin{align}
$e_i=W^Tv_i$
% \end{align} 
where $v_i$ is a one-hot vector for the feature value $f_i$, $W \in \mathbb{R}^{|F|\times D}$ refers to the embedding table
% , $|F|$ refers to the vocabulary size, $D$ is pre-defined embedding dimension 
and $e_i \in \mathbb{R}^D$ is the embedding vector of $f_i$.
% Since the dimension $D$ is pre-defined, all of the feature value in this category have the same embedding dimension, which neglects the different frequency among feature values leading to a sub-optimized embedding vectors.
Then we define a mask vector $m_i \in \{0,1\}^D$ for $f_i$.
This mask vector should satisfy
\begin{align}
m_{i,j}=\left\{\begin{array}{lc}1&j\leq k_i\\0&j > k_i\end{array}\right.
\label{Equ:basic idea mij}
\end{align} 
where $k_i \in [0,D-1]$ is a learnable integer parameter which is influenced by the frequency of $f_i$.
Then, to allow different $f_i$ can adjust its embedding dimension, the basic idea is that we can use the mask vector $m_i$ to mask the embedding vector $e_i$, i.e., 
% \begin{align}
$\hat{e}_i=m_i \odot e_i$
% \end{align}
where $\odot$ represents the element-wise multiply.
Since the value whose index is larger than $k_i$ in $\hat{e}_i$ is zero, the masked embedding vector $\hat{e}_i$ can be taken as an embedding vector where the embedding dimension is adaptively adjusted by the mask vector, and the first $k_i+1$ dimensions (i.e., from 0-th to $k_i$-th dimension) of $e_i$ is selected.

% Hear, we highlight some nice properties of this dear
\noindent\textbf{Memory Saving.}
When storing $\hat{e}_i$, we can simply drop the zero values in $\hat{e}_i$ to save memory and when fetching the stored vector, we can simply re-pad zero values to recover $\hat{e}_i$.
% (2) \textbf{Warm Start.}
% Such idea is friendly to the warm start of embedding vectors, since $e_i$ can not only be initialized randomly, but also be initialized from existing t

\noindent\textbf{Embedding Usage.}
When embedding vectors are assigned with different dimensions, all of the existing methods \cite{ginart2019mixed,zhao2020memory,zhao2020autoemb,joglekar2020neural,liu2020automated} have to design an additional layer to unify these vectors to a same length to fit the following uniform MLP layers in DLRMs.
Unlike these methods, our method does not need any additional layers since the masked embedding vectors $\hat{e}_i$ have the same length by zeros paddings, and can be directly fed into the following layers.

\noindent\textbf{Why select first $k_i+1$ dimensions?}
% We should point out that, there are many strategies to select $k_i+1$ dimension of $e_i$ to reserve, e.g., selecting first $k_i+1$ or last $k_i+1$ dimension.
In this paper, We take the strategy about selecting first $k_i+1$ dimensions (i.e., from 0-th to $k_i$-th dimension) of $e_i$ as an example  and others (e.g.,  selecting last $k_i+1$ dimensions) are also allowed.
One should keep in mind is that the select strategy should follow some rules.
In other words, randomly selecting $k_i+1$ dimensions of $e_i$ is not a good strategy because we can hardly directly drop and recover these zeros values in $\hat{e}_i$ to save memory due to the random distribution of zeros values in $\hat{e}_i$, and it also prevents the model to characterize each feature value since the same feature value may be mapped to different embedding vectors due to the random selection.

% In this way, we dynamically assign different embedding dimensions to different feature values, allowing the feature value with high-frequency to obtain more representation dimension and the feature value with low-frequency to avoid a high embedding dimension.
% The next question is that how to generate such mask vectors $m_i$.
% The following section presents the proposed Adaptively-Gated Twins-of-Router Layer for the mask vector generation.

\begin{figure}[t]
\centering
\includegraphics[width = .4\textwidth]{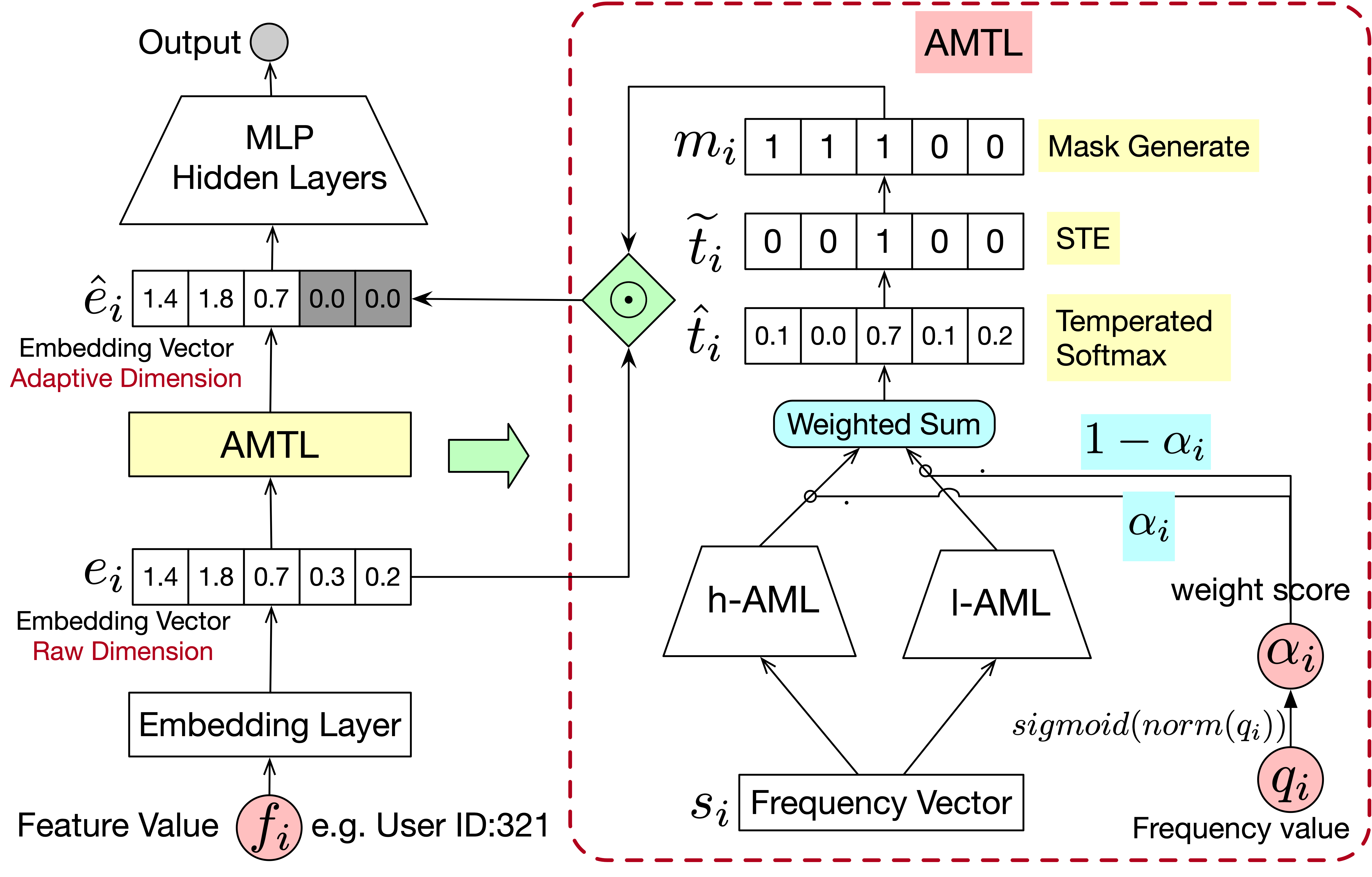}
\vspace{-1em}
\caption{The framework of AMTL.
}
\vspace{-2em}
\label{figure:Adaptively-Gated Twins-Of-Router Layer}
\end{figure}

\subsection{Adaptively-Masked Twins-based Layer}
\label{sec:Adaptively-Gated Twins-Of-Router Layer}
Adaptively-Masked Twins-based Layer (AMTL) is designed to generate a mask vector $m_i$ in Eq \ref{Equ:basic idea mij} for each feature value $f_i$.
The framework of AMTL is shown in Fig \ref{figure:Adaptively-Gated Twins-Of-Router Layer}.
% Then what are the input and the output of the AMTL?

\subsubsection{Input and Output}
\textbf{Input:} 
Since $m_i$  is required to be adjusted by the feature frequency, to allow AMTL to have such frequency knowledge, we take the frequency attribute (e.g., the appear times in history, the frequency rank in this feature field and so no) of $f_i$ as the input of AMTL.
The input sample is denoted as $s_i \in R^z$ and $z$ is the input dimension.
% For example, we can map the frequency value $q_i$ of $f_i$  into an embedding vector and take this vector as the input 
% (Note $q_i$ refers that $f_i$ appears $q_i$ times in history).
\textbf{Output:} 
% According to Eq \ref{Equ:basic idea mij}, the mask vector $m_i$ is determined by the value $k_i$. 
The output of AMTL is a one-hot vector (called selection vector) to represent $k_i$ in Eq \ref{Equ:basic idea mij}.

\subsubsection{Architecture}
We propose a twins-based (i.e., two branches) architecture and each branch is an Adaptively-Masked Layer (AML). 
% (The motivation of such designed is presented in Section xx).
Note parameters of the two branches are not sharing.
Both of AML is a multilayer perceptron:
% \begin{align}
$h_l=\sigma(W_l^T\;h_{l-1}+b_l)$
% \end{align}
 where $h_0$ is the frequency vector, $W_l$ and $b_l$ are the parameters of the $l$-th layer, $\sigma$ is an activation function and $h_L \in \mathbb{R}^D$ is the output of the last layer.

The motivation of such twins design is that if we only take a single branch (i.e., AML), the parameters update of AML will be dominated by the high-frequency feature values due to the unbalanced problem. 
Specifically, since high-frequency feature values appear more times in samples, the major part of the input sample of AML represents high-frequency vectors.
Then, the parameters of AML may be heavily influenced by the high-frequency vectors and AML may blindly select large embedding dimensions.
 % no matter the input vector refers to high-frequency or low-frequency.
 % which is opposite to the requirement giving fair dimension decision among inputs.
Hence we design twins-based architecture to address this problem where the two branches (i.e., \textit{h-AML} and \textit{l-AML}) are used for high- and low- frequency samples respectively.
In this way, the parameters of the \textit{l-AML} will not be dominated by high-frequency samples and can give an unbiased decision.

\noindent\textbf{Weighted Sum.}
However, one challenge is that we can hardly give a threshold to differentiate the high- and low- frequency samples to feed different samples to different branches.
Hence, we propose a soft decision strategy.
Specifically, 
% for feature value $f_i$,
 % we define the corresponding sample of AMTL as $s_i \in \mathbb{R}^K$. 
% Besides, 
we define the frequency value of $f_i$ as $q_i$ which refers to the present times of $f_i$ in history,
% (e.g., if the value $f_i$ is present $q_i$ times in the history, the frequency value is set $q_i$).
and feed the input sample $s_i$ into \textit{h-AML} and \textit{l-AML} respectively.
A weighted sum is applied on the $L$-th outputs (i.e., $h_L^{(\textit{h-AML})} \in \mathbb{R}^D$ and $h_L^{(\textit{l-AML})} \in \mathbb{R}^D$ ) of \textit{h-AML} and \textit{l-AML}, i.e., 
  % are expressed as $h_L^{(\textit{h-AML})} \in \mathbb{R}^D$ and $h_L^{(\textit{l-AML})} \in \mathbb{R}^D$ respectively.
% Then a weighted sum is applied on these two outputs, i.e.,
\begin{align}
h_L^{(\textit{AMTL})} &= \alpha_i*h_L^{(\textit{h-AML})} + (1-\alpha_i)*h_L^{(\textit{l-AML})} \\
\alpha_i &= sigmoid(norm(q_i))
\end{align}
where $\alpha_i \in [0,1]$ is the weight score which is influenced by $q_i$.
$norm$ operation normalizes $q_i$ to a standard normal distribution which allows $\alpha_i$ is distributed smoothly around $1/2$.
Otherwise, all $\alpha_i$ may be close to one without the $norm$ operation.
In this way, for the samples with high-frequency, the corresponding $h_L^{(\textit{AMTL})}$ is dominated by $h_L^{(\textit{h-AML})}$ due to a large $\alpha_i$.
Then the parameters of \textit{h-AML} are mainly updated during back-propagation and vice versa.
Hence, AMTL can adjust the gradients of \textit{h-AML} and \textit{l-AML} to address the unbalanced problem.
Note here we only give an example to calculate the weight value $\alpha_i$, other ways are also allowed as long as the produced $\alpha_i$ has similar properties.

% \subsection{Adaptively-Gated Router Layer}
% \label{sec:Adaptively-Gated Router Layer}
% Adaptively-Gated Router Layer (AML) is designed to generate a mask vector $m_i$ for each feature value $f_i$.
% The framework of AML is shown in Fig \ref{figure:Adaptively-Gated Router Layer.}.
% Then what are the input and the output of the AML?

% \textbf{Input:} Since the mask vector is required to be adjusted by the frequency of a feature value, to allow AML to be aware of such frequency knowledge, we take the frequency attribute of this feature value as the input of AML.
% An example for the frequency vector can be obtained by mapping the frequency value $q_i$ of $f_i$  into an embedding vector 
% (Note if the value $f_i$ is present $q_i$ times in the history, the frequency value is set $q_i$).

% \textbf{Output:} According to Eq \ref{Equ:basic idea mij}, the mask vector $m_i$ is determined by the value $k_i$. 
% Therefore the output of AML is a one-hot vector (called selection vector) to represented this integer $k_i$.

% To connect the input with the output of AML, the architecture of AML is designed as follows (see Fig \ref{figure:Adaptively-Gated Router Layer.}), where we take multilayer perceptron (MLP) as the hidden layer:
% \begin{align}
% h_i&=\sigma(W_i^T\;h_{i-1}+b_i) 
% \end{align} where $h_0$ is the frequency vector, $W_i$ and $b_i$ are the parameters, $\sigma$ is a activation function and $h_L \in \mathbb{R}^D$.

Then we apply softmax function on $h_L^{(\textit{AMTL})}$, i.e.,
% \begin{align}
% o_{i,p}=\frac{exp(h_{L,p})}{{\displaystyle\sum_j}exp(h_{L,j})}
% \end{align} 
\begin{align}
o_{i,p}=exp(h_{L,p}^{(\textit{AMTL})}) / {\textstyle\sum_j}exp(h_{L,j}^{(\textit{AMTL})})
\end{align} 
where $o_i \in \mathbb{R}^D$ refers the probability to select different embedding dimension of $f_i$, and  $o_{i,p}$ is the $p$-th element of $o_i$.
The selection vector can be obtained by 
% \begin{align}
% \label{equ:one_hot}
% t_i=one\_hot(\underset{p}{argmax}(o_{i,p}))
% \end{align} 
\begin{align}
\label{equ:one_hot}
t_i=one\_hot(argmax_p(o_{i,p}))
\end{align} 
Then the corresponding mask vector can be generated by
\begin{align}
m_i&=M^Tt_i
\end{align}
where $M \in \mathbb{R}^{D\times D}$ is a pre-defined mask matrix and $M_{i,j}=1$ when $j\leq i$ otherwise $M_{i,j}=0$. 
% Note, we take $t_i$ as the one-hot form of the $k_i$ introduced in Eq \ref{Equ:basic idea mij}.
Then the masked embedding $\hat{e}_i$  can be obtained by $m_i$.
% which will be feed into the following layers of DLRM in an end-to-end manner.
% Finally, AMTL can be joined trained with DLRM.
Note in practice, we usually apply different AMTLs on different important feature fields (e.g., User ID and Item ID) and the parameters of these layers are not sharing for the purpose of field awareness.

\subsubsection{Relaxation}
\label{sec:Continues Relaxation}
However, the problem is that the learning process of AMTL is non-differentiable due to the discrete process in Eq \ref{equ:one_hot}.
It means the parameters of AMTL cannot be directly optimized by a stochastic gradient descent (SGD).
To address this problem, we relax $t_i$ to a continuous space by temperated softmax \cite{hinton2015distilling,jang2016categorical,maddison2016concrete}.
Concretely, the $p$-th element of $t_i$ can be approximated as
% \begin{align}
% \label{equ: approx t}
% t_{i,p} \approx \hat{t}_{i,p}=\frac{exp(h_{L,p}/T)}{{\displaystyle\sum_j}exp(h_{L,j}/T)}
% \end{align}
\begin{align}
\label{equ: approx t}
t_{i,p} \approx \hat{t}_{i,p}=exp(h_{L,p}^{(AMTL)}/T)/{{\textstyle\sum_j}(exp(h_{L,j}^{(AMTL)}/T)}
\end{align}
where $T$ is the temperature hyper-parameter.  
When $T \rightarrow 0$, this approximation becomes exact.
Since $\hat{t}_i$ is a continuous vector with differentiable process, SGD can be naturally applied.
Hence, instead of learning the discrete vector $t_i$, we learn $\hat{t_i}$ to approximate $t_i$. 

% Although the non-differentiable problem can be well settled by continuous relaxation during training,
However, there exists an information gap between training and inference phases when using temperature softmax.
Specifically, we use the vector $\hat{t}_i$ for training. 
While in inference, we only use the discrete vector $t_i$.
To close this gap, inspired by the idea Straight-Through Estimator (STE) \cite{bengio2013estimating}, we rewrite $t_i$ as
\begin{align}
\widetilde{t}_i = \hat{t}_i + stop\_gradient(t_i-\hat{t}_i)
\end{align}
where stop\_gradient is used to prevent the gradient from back-propagation through it.
Since the forward pass is not affected by stop\_gradient, $\widetilde{t}_i=t_i$ during this phase.
 % which equal to set $T \rightarrow 0$ in the forward pass.
For the back-propagation, it avoids the non-differentiable process by stop\_gradient.

\section{Experiments}

% \begin{table}[t]
% \caption{The statistic of datasets}
% \begin{tabular}{|l|c|c|c|}
% \hline
%         & \#Data    & \#User ID   & \#Item ID   \\ \hline
% MovieLen & 1,000,209 &  6,040 & 3,706  \\ \hline
% IJCAI-AAC & 478,138 & 197,694 & 10,075 \\ \hline
% Taobao & 50 billion  & 100 million & 80 million\\ \hline

% \end{tabular}
% \label{table:The statistic of datasets}
% \end{table}

\subsection{Experiment Setup}
\noindent\textbf{Data Sets.}
% The used datasets are summarized as follows:
(1) MovieLens \footnote{https://grouplens.org/datasets/movielens/} is a user review data about movies and is collected from MovieLens website.
% Each rate is ranges from 1 to 5.
There are a total of 1,000,209 records.
% of approximately 3,706 movies made by 6,040 users.
% To meet the CTR prediction setting,
% For simplicity, we take the rate greater than 3 as a positive rate and others as a negative rate.
% Then, we need to predict whether the user gives a positive or negative rate to this movie.
(2) IJCAI-AAC \footnote{https://tianchi.aliyun.com/competition/entrance/231647/information} is  collected from a sponsored search in E-commerce.
% Each record refers to whether a user purchases the displayed item after clicking this item.
% Such kind of dataset can be easily used in CTR prediction tasks.
There are a total of 478,138 records.
% , and 197,694 users and 10,075 items.
(3) Taobao Dataset is an industrial dataset which is constructed from Taobao.
 % the largest online retail platform in China.
% , the largest online retail platform in China.
% Each record refers to that an item is browsed by an user.
% The label is defined as whether the user clicks this item.
There are a total of 50 billion around records.

% The statistics of the data sets are summarized in Table \ref{table:The statistic of datasets}.

\noindent\textbf{Baselines.}
We consider different kinds of state-of-the-art embedding methods as baselines
% \begin{itemize}
% \item 
(1) Standard: traditional Fixed-based Embedding (FBE).
 % is a standard embedding method where each feature value in the same feature field is assigned with a fixed embedding dimension.
% \item 
(2) Rule-Based: MDE \cite{ginart2019mixed} divides different feature values into several blocks by their frequency, and assigns different embedding dimensions for different blocks by rules.
% This method is chosen to show the performance of rule-based methods.
% \item 
(3) NAS-Based: AutoEmb \cite{zhao2020autoemb} adopts NAS to select embedding dimensions among some candidate embedding dimensions for different feature values.
% We chose this method to evaluate the performance of the NAS-Based method.

% \item \textbf{Random-based Embedding (RBE)} assigns different embedding lengths for different feature value randomly, i.e., generating the selection vector $t_i$ randomly. 
% We take this method to evaluate whether the learned select vector $t_i$ by AMTL is useful.

% \begin{table}[t]
% \centering
% \caption{The Results about CTR Tasks}
% \vspace{-1em}
% \begin{adjustbox}{max width=\linewidth}
% \begin{tabular}{lccc}
%  \toprule 
%    AUC(\%)     &IJCAI-AAC& MovieLens& Taobao \\ \hline
% FBE     &62.37&80.84& 73.64   \\ 
% MDE     &62.4&80.51& 73.54   \\ 
% AutoEmb &62.85&81.28& 73.67   \\ \hline 
% % RBE     &62.48&81.01& 73.68        \\ \hline \hline
% AMTL    &\textbf{63.45}&\textbf{81.50}& \textbf{74.02}   \\ 
% \bottomrule
% \end{tabular}
% \end{adjustbox}
% \vspace{-1em}
% \label{tabel:The Results about Click Prediction Tasks}
% \end{table}

\begin{table}[t]
\centering
\caption{The results about CTR tasks.}
\vspace{-1em}
\begin{adjustbox}{max width=\linewidth}
\begin{tabular}{lcccc}
 \toprule 
   AUC(\%)     &FBE& MDE& AutoEmb &AMTL \\ \hline
IJCAI-AAC     &62.37&62.40& 62.85&\textbf{63.45}   \\ 
MovieLens   &80.84&80.51& 81.28&\textbf{81.50}   \\ 
Taobao &73.64&73.54&73.67&\textbf{74.02}   \\ 
% RBE     &62.48&81.01& 73.68        \\ \hline \hline
\bottomrule
\end{tabular}
\end{adjustbox}
\vspace{-1em}
\label{tabel:The Results about Click Prediction Tasks}
\end{table}

\noindent\textbf{Setting.}
% In this section, we  introduce the training details.
% To have a fair comparison, for each feature field in different datasets, 
The maximal embedding dimension and the DLRM bone-skeleton of all methods are set the same.
% There are many categorical feature fields.
The dimension selection strategy is applied to the feature fields which are related to the user and item property (e.g., User ID and Item ID).
% Fairly, all the methods focus on the same kinds feature fields.
% For Taobao data, we apply the dimension selection on the category feature "user id" as an example for simplicity.
% The maximal embedding length of "user id" is set 300.
For AutoEmb, the candidate dimension list is set smoothly by following the original paper \cite{zhao2020autoemb}.
% For example, the candidate dimensions of "User ID" in Taobao are \{10,50,150,300\}.
% the corresponding candidate or block embedding list is [10,50,150,300].
% The frequency vector is obtained by bucking the frequency value and mapping it into an embedding vector.
The temperature $T$ in Eq \ref{equ: approx t} is set by grid search.
 % over \{0.2,0.4,0.6,0.8,1.0\}.
% The optimizer is Adagrad with a learning rate of 0.005. 
% The batch size is 1024 for all datasets.

\begin{table}[t] \smaller
\caption{Memory cost comparison.}
\vspace{-1em}
\begin{tabular}{lcccc}
\toprule 
Methods    & FBE & MDE & AutoEmb & AMTL\\ \hline
Avg(Dim)     & 300   &170 & 206 & 110\\ 
Ratio    & 100\%   &56.7\% & 68\% & 36.7\% \\  
\bottomrule
\end{tabular}
\vspace{-1em}
\label{tabel:Memory Cost Comparison}
\end{table}

\subsection{Click-Through Rate Prediction Tasks}
\label{sec:Click Prediction Task}
Here, we compare our method AMTL with baselines on click-through rate (CTR) prediction tasks and take the AUC \cite{fawcett2006introduction} score as the metric. 
Note a slightly higher AUC at \textbf{0.1\%-level} is regarded as significant for the CTR task \cite{zhou2018deepdin,autoint}.
As shown in Table \ref{tabel:The Results about Click Prediction Tasks}, we can conclude that:
(1) Compared with FBE, AMTL can archive better performance in all datasets.
It shows that adopting an unfixed embedding dimension can improve the model performance.
% (2) Compared with dimension selection method (including MDE and AutoEmb), AMTL
(2) Compared with the rule-based method (i.e., MDE), AMTL outperforms MDE.
Besides, MDE only obtains similar performance with FBE.
It indicates a rough human rule on dimension selection cannot always guarantee an improvement.
% Hence, AMTL is a more wise strategy to adaptively select embedding dimensions for different feature values.
(3) For the NAS-based method (AutoEmb), AMTL also archives better performance.
It demonstrates that AMTL adopts a more suitable scheme i.e., selecting a dimension from a continuous integer space.
% The possible reasons can be summarized as 1) AMTL allows the feature value to select a dimension from a continuous integer space rather than a restricted set of predefined dimensions in AutoEmb. 
% 2) The NAS process needs more effort to be optimized \cite{zhao2020autoemb,joglekar2020neural,joglekar2020neural} which may bring unstable influence on model training and inference. 

% (2) Compared with other dimension selection methods including AutoEmb and RBE, AMTL also outperforms these baselines.
% One of the possible reasons is that AMTL allows the feature value to select dimension from a continuous space while the dimension of baselines are restricted in a set of predefined dimensions. 
% The improvement demonstrates that the idea of mask is a wiser way to adjust dimension for features.
% Furthermore, considering the flexibility of AMTL (see Section \ref{sec:Introduction}), such idea brings a new and good direction for the dimension selection problem. 
% (3) We also note that although MDE use an unfixed embedding dimension, MDE is still worse than FBE.
% It shows a rough design for assigning embedding dimension may bring negative impacts on models.

% \begin{table}[t]
% \caption{Memory Cost Comparison.}
% \begin{tabular}{lcc}
% \toprule 
% Methods    & Ave(Dim) & Ratio\\ \hline
% FBE     & 300   &100\%\\ 
% MDE     & 170   &56.7\%\\ 
% AutoEmb & 206   &68.7\%\\ \hline \hline
% AMTL    & 110   &36.7\% \\ 
% \bottomrule
% \end{tabular}
% \label{tabel:Memory Cost Comparison}
% \end{table}

\subsection{Memory Cost Comparison}
\label{sec:Memory Cost Comparison}
% As discussed in Section \ref{sec:Introduction}, the traditional methods adopt a fixed and uniform embedding dimension which may cost a lot of memory.
% Hence, 
Here, we compare the memory cost of different methods.
Since the memory size of the embedding matrix is in direct proportion to the dimension \cite{zhang2020model,shi2020compositional}, for simplicity, the averaged dimension (i.e., Avg(Dim)) are reported.
% There are various feature fields in different datasets, 
We take the feature field "User ID" in Taobao as an example (others can have similar conclusions), and its' maximal dimension is set as 300.   
Table \ref{tabel:Memory Cost Comparison} shows the results. 
We also show the Avg(Dim) ratio compared with FBE.
We can find that (1) Compared with FBE, all the dimension selection methods can save memory size by reducing the embedding to a suitable dimension.
(2) Since AMTL allows a more flexible dimension selection in a continuous integer space, 
% it is more significant to reduce memory cost and decrease the memory usage 
it reduces memory cost more significantly by around 60 \%.

% \begin{table}[t]
% \caption{The Results of Warm Start}
% \vspace{-1em}
% \begin{tabular}{lcc}
% \toprule 
%    AUC(\%)     &Warm Start & Random Start \\ \hline

% MDE    &75.43& 73.54   \\ 
% AutoEmb &75.62& 73.67   \\ \hline 
% % RBE     &62.48&81.01& 73.68        \\ \hline \hline
% AMTL    &\textbf{77.22}& \textbf{73.82}   \\ 
% \bottomrule
% \end{tabular}
% \vspace{-1em}
% \label{tabel:The Results on the Dataset Taobao with Warm Start}
% \end{table}

\begin{table}[t]\smaller
\caption{The results of warm start.}
\vspace{-1em}
\begin{tabular}{lcccc}
\toprule 
   AUC(\%)     &FBE&MDE & AutoEmb&AMTL \\ \hline

Warm Start    &77.05&75.43& 75.62 & \textbf{77.22}  \\ 
Random Start &73.64&73.54& 73.67  & \textbf{74.02} \\ 
\bottomrule
\end{tabular}
\vspace{-1em}
\label{tabel:The Results on the Dataset Taobao with Warm Start}
\end{table}

\begin{table}[t]\smaller
\caption{The results of ablation study.}
\vspace{-1em}
\begin{tabular}{lcccc}
\toprule 
    & FBE   & AMTL & AML& AMTL-nSTE \\ \hline
AUC(\%) & 62.37 & 63.45 & 62.97& 63.01     \\ 
\bottomrule
\end{tabular}
\vspace{-1.5em}
\label{table:The results of ablation study}
\end{table}

\subsection{Evaluation for the Warm Start}
\label{sec:Effectiveness for the Warm Start of the Embedding Matrix}
% As discussed in Section \ref{sec:Introduction}, one of the major problems of existing methods (including rule-based and NAS-based methods) is that they cannot make full use of the well-trained embedding matrix in existing DLRM.
% In other words, if adopting various embedding dimensions for each feature field, they have to drop the fully trained embedding matrix of DLRM which is severing on the real systems, and initialize the new embedding matrix from a random start.
% As for AMTL, there is no need to drop the well-trained embedding matrix, and it gives a convenient scheme by generating a masking vector for each embedding vector.
% Hence, AMTL is more friendly to the warm start of the embedding matrix.

% \subsubsection{Setting}
Here, we conduct experiments to evaluate the warn start on the dataset Taobao which is close to the industrial and real system.
% For convenience, we denoted the dimension selection based DLRM as AMTL-DLRM, MDE-DLRM, AutoEmb-DLRM, and defines the model which has been fully trained and are severing online as online-DLRM.
% Note the architecture of these models is setting the same.
There are two kinds of parameters in DLRM, i.e., the parameters of embedding matrix and hidden layers.
In the warm start setting, for existing dimension selection methods, we initialize the parameters of hidden layers by loading the parameters from the online model in Taobao, and the parameters of embedding matrix are randomly initialized due to the inability on the warm start of embedding matrix.
While, since AMTL and FBE can warm start the parameters both of the embedding matrix and the hidden layers, we load both of them from the online model for these two methods.
The results are shown in Table \ref{tabel:The Results on the Dataset Taobao with Warm Start}.
The results of random start are also provided.
We can find that 
% (1) In the warm start setting, all of the methods can make great progress. 
% It indicates that making use of the parameters in online model can archive a remarkable improvement.
% (2) 
compared with MDE and AutoEmb, AMTL can perform better in the warm start setting.
Specifically, compared with the best baseline AutoEmb, the gain of AMTL is 1.6\% in the warm start manner, which is 5$\times$ times larger than the gain in the random start manner.
Furthermore, due to the inability for the warm start of the embedding matrix, MDE and AutoEmb even perform worse than the standard full embedding.
It demonstrates that the dimension selection scheme designed in AMTL is a more wise and flexible way in real application systems.

\subsection{Evaluation for Frequency}
\label{sec:Dimension Selection Analysis.}
\begin{figure}[t]
\vspace{-1em}
  \centering
    \subfigure[AMTL]{
    \includegraphics[width= .2\textwidth]{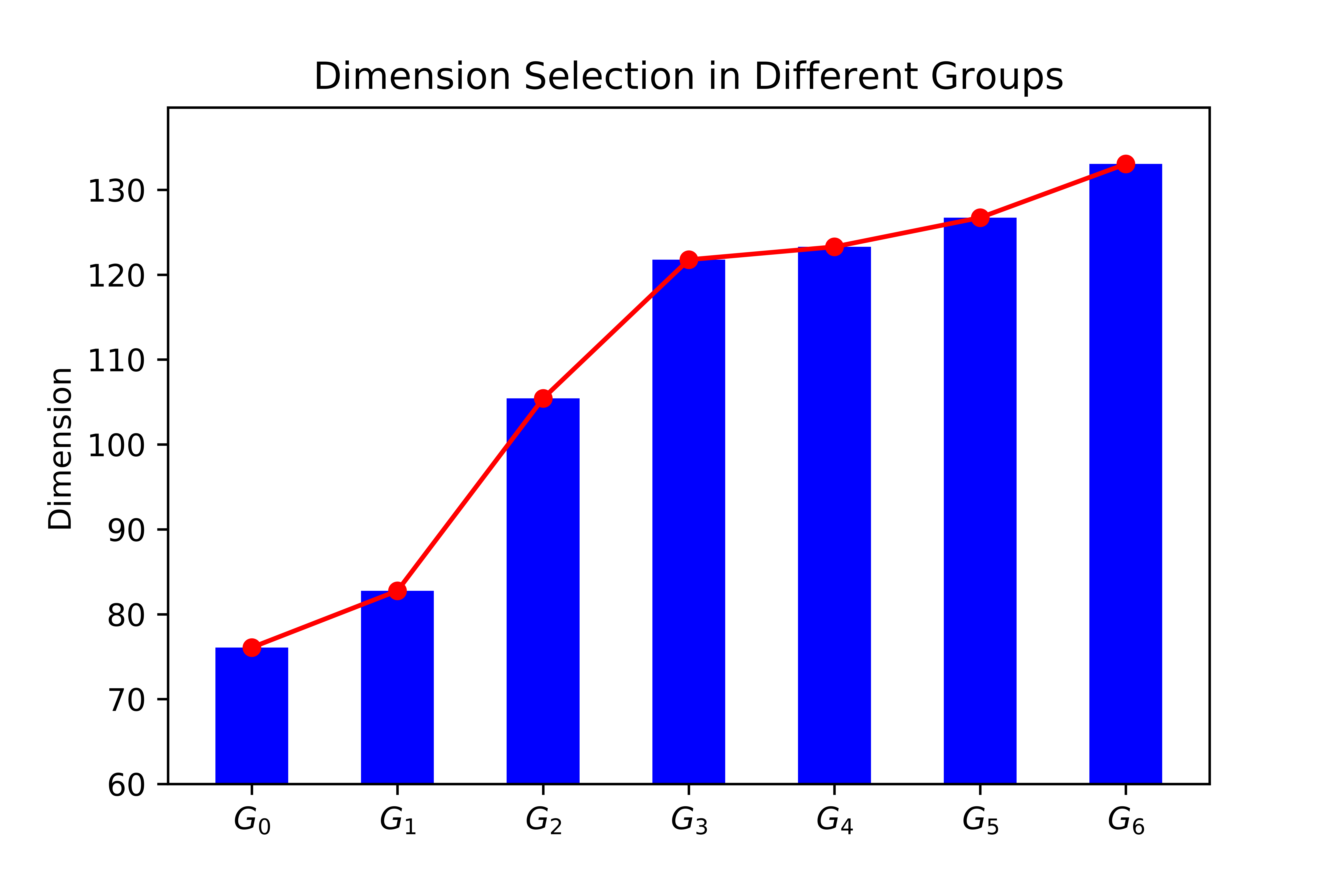}
    \label{fig:subfigure2}
  }
      \subfigure[AML]{
    \includegraphics[width= .2\textwidth]{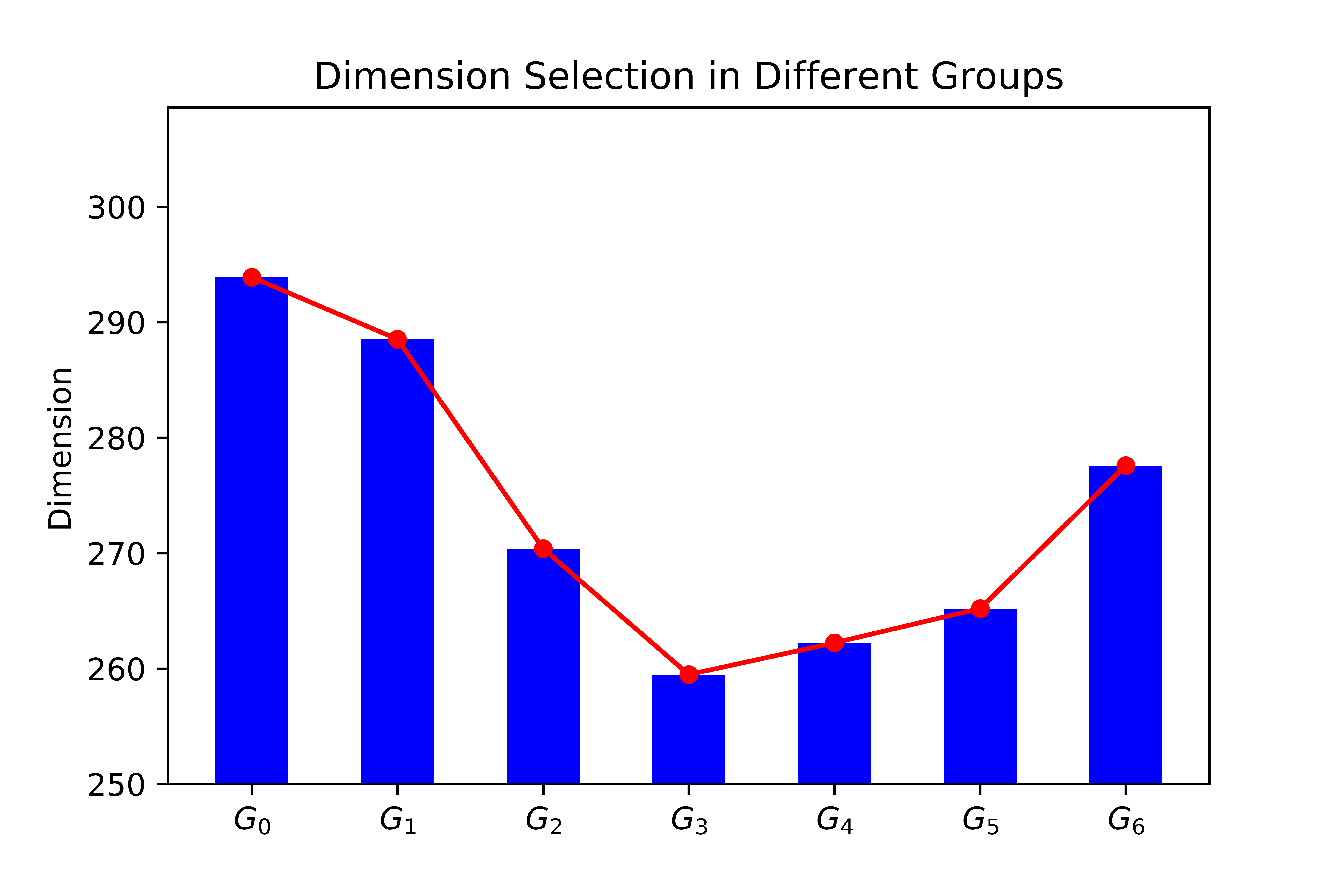}
    \label{fig:subfigure2}
  }
\vspace{-1.5em}
\caption{Dimension section of AMTL and AML.}
\vspace{-1em}
\label{figure:Dimension Section Comparison between AMTL and AML in Different Groups}
\end{figure}

% As reported in Section \ref{sec:Click Prediction Task}, AMTL archives the best results overall cases.
% In this section, we give a deeper analysis to discuss why AMTL works.
% As we know, the motivation of AMTL is to allow the low-frequency feature values to have small embedding dimensions to avoid over-fitting and give a large embedding dimension to high-frequency feature values to represent themselves sufficiently.
Here we analyze whether AMTL can give suitable dimensions for different feature values.
Specifically, we divide the feature value into 7 groups (i.e., $G_i, i \in \{0,1,2,...,6\}$) by frequency and the average frequency of different groups is increased from $G_0$ to $G_6$.
% Then, we evaluate AMTL from the perspective of dimension selection and AUC in different groups.
Note there are too many feature fields in different datasets, we take the feature field "User ID" in Taobao as an example, and others have similar results.
% \begin{figure}[t]
% \centering
% \includegraphics[width = .4\textwidth]{dimension_selection.png}
% \caption{Dimension Selection in Different Groups
% }
% \label{figure:Dimension Selection Analysis.}
% \end{figure}
% \subsubsection{Dimension Selection Analysis.}
% \label{sec:Dimension Selection Analysis.}
% As we know, the motivation of AMTL is to allow the low-frequency feature values to have small embedding dimensions to avoid over-fitting and give a large embedding dimension to high-frequency feature values to represent themselves sufficiently.
% Therefore, we analyze the dimension selection for different feature values.
% Specifically,
The averaged embedding dimension in different groups is reported in Fig \ref{figure:Dimension Section Comparison between AMTL and AML in Different Groups} (a).
% Some observations are summarized as follows:
% (1) 
It shows that when the frequency increases (i.e., from $G_0$ to $G_6$), the selected average dimension is increased. 
It indicates AMTL can assign suitable embedding dimensions to different frequency feature values adaptively.

\subsection{Ablation Study}
Here, we conduct an ablation study on the twins-based architecture and STE.
Due to the limited space, the results on IJCAI-AAC are reported and similar conclusions can be found from other datasets.

\noindent\textbf{Evolution on twins-based architecture.}
\label{sec:Evolution on two branches architecture}
We compared the AUC score between AMTL and AML (only a single branch) on CTR task.
From Table \ref{table:The results of ablation study}, although both AML and AMTL perform better than FBE, AMTL archives a higher performance. 
% Besides, AML only archives similar results with FBE.
It indicates twins-based architecture plays an important role in feature learning.
Besides, similar to Section \ref{sec:Dimension Selection Analysis.}, we also visualize the dimension selection of AML in Fig \ref{figure:Dimension Section Comparison between AMTL and AML in Different Groups} (b). 
We can find that AML only successfully gives suitable dimensions in high-frequency groups (i.e., $G_3$ to $G_6$).
In low-frequency groups, due to the unbalanced problem, it blindly gives high dimensions for low-frequency values.
And the lower the frequency, the worse it is.

\noindent\textbf{Evolution on STE.}
% As introduced in Section \ref{sec:Continues Relaxation}, STE is introduced to tackle the information gap between training and inference phases.
Here, we analyze the effectiveness of STE, and implement a variant of AMTL without STE, denoted as AMTL-nSTE.
% Similarly like Section \ref{sec:Evolution on two branches architecture}, we also report the AUC score on dataset IJCAI-AAC for the same reasons.
% The results about AUC score on dataset IJCAI-AAC are presented in Table \ref{table:The results of ablation study}.
% Similarly, we also provide the results of FBE for comparison.
From Table \ref{table:The results of ablation study}, compared with AMTL-nSTE, AMTL can archive better performance.
It demonstrates the usefulness to bridge the information gap between training and inference phases by STE.

\subsection{Time Cost Analysis}
In this section, we conduct experiments to report the time cost per epoch of different methods on IJCAI-AAC (similar conclusions can be found in other datasets).
As shown in Table \ref{table:The Time Cost of Different Methods}, we find that compared with FBE, the dimension selection methods need more time to train the model per epoch due to the dimension selection processes.
During inference, we can directly look up the learned embedding table which has adaptive dimensions without the process of dimension selection to save time.

\begin{table}[t]
\caption{The time cost of different methods}
\vspace{-1em}
\begin{tabular}{lcccc}
\toprule 
    & FBE   & MDE &AutoEmb  & AMTL \\ \hline
per epoch (s) & 13.3 & 13.8 &15.7 &14.5     \\ 
\bottomrule
\end{tabular}
\vspace{-2em}
\label{table:The Time Cost of Different Methods}
\end{table}

\section{Conclusion}
Traditional embedding learning methods usually adopt a fixed dimension for all features which may cause problems in space complexity and performance.
To address this problem, we propose a novel dimension selection method called AMTL which produces a mask vector to mask the undesired dimensions for different feature values.
% In this way, the memory size of embedding matrix can be reduced and the requirements for different dimensions can be satisfied.
% Furthermore, such a gated layer based design brings a flexibility in dimension selection and applications.
Experimental results show that the proposed method can archive the best performance on all tasks especially in the case of embedding warm start, give a suitable dimension for different features and save memories at the same time.

% \bibliographystyle{ACM-Reference-Format}


\end{document}